\pdfoutput=1

\documentclass[11pt]{article}

\usepackage[final]{acl}

\usepackage{times}
\usepackage{latexsym}
\usepackage{enumitem}
\usepackage{amsmath}
\usepackage{amssymb}
\usepackage{cleveref}

\usepackage[T1]{fontenc}

\usepackage[utf8]{inputenc}

\usepackage{microtype}

\usepackage{inconsolata}

\usepackage{graphicx}

\usepackage{tcolorbox}

\newcommand{\model}{\mathcal{M}}
\newcommand{\inputsentence}{x}

\newcommand{\embed}{\mathrm{Emb}}
\newcommand{\E}{\mathrm{E}}
\newcommand{\R}{\mathbb{R}}
\newcommand{\activation}{a}

\title{Auxiliary Metrics Help Decoding Skill Neurons in the Wild}

\author{
  Yixiu Zhao$^{*}$ \quad
  Xiaozhi Wang\thanks{ indicates equal contribution.} \quad
  Zijun Yao \quad
  Lei Hou \quad
  Juanzi Li \\
  Tsinghua University, Beijing, China, 100084 \\
  \texttt{yx-zhao22@mails.tsinghua.edu.cn} \\
}

\begin{document}

\tcbset{
  colframe=blue!75!black, 
  colback=blue!10,        
  boxrule=1pt,            
  sharp corners,          
  fonttitle=\bfseries,    
}

\maketitle
\begin{abstract}
Large language models (LLMs) exhibit remarkable capabilities across a wide range of tasks, yet their internal mechanisms remain largely opaque. In this paper, we introduce a simple, lightweight, and broadly applicable method with a focus on isolating neurons that encode specific skills. Building upon prior work that identified “skill neurons” via soft prompt training on classification tasks, our approach extends the analysis to complex scenarios involving multiple skills. We correlate neuron activations with auxiliary metrics—such as external labels and the model’s own confidence score, thereby uncovering interpretable and task-specific behaviors without the need for manual token aggregation. We empirically validate our method on tasks spanning open-ended text generation and natural language inference, demonstrating its ability to detect neurons that not only drive known skills but also reveal previously unidentified shortcuts in arithmetic reasoning on BigBench. 
\end{abstract}

\section{Introduction}



Large language models (LLMs) have become increasingly powerful. As these models grow in complexity, interpretability research becomes essential—to understand their inner workings and also steer them safely~\citep{bereska2024mechanisticinterpretabilityaisafety}.

Interpretability research typically unfolds in two stages \citep{räuker2023transparentaisurveyinterpreting}. In the \textit{observation} stage, researchers attribute specific model behaviors to substructures within the network. For instance, prior studies have shown that certain neurons encode specific skills, like sentiment detection or fact retrieval \citep{radford2017learninggeneratereviewsdiscovering,gurnee2023findingneuronshaystackcase,wang2022findingskillneuronspretrained,bills2023language,choi2024automatic}. In the \textit{intervention} stage, these insights are taken further by modifying the weights or activations of the identified substructures, thereby causally influencing the model’s behavior.

\begin{figure}
    \centering
    \includegraphics[width=\linewidth]{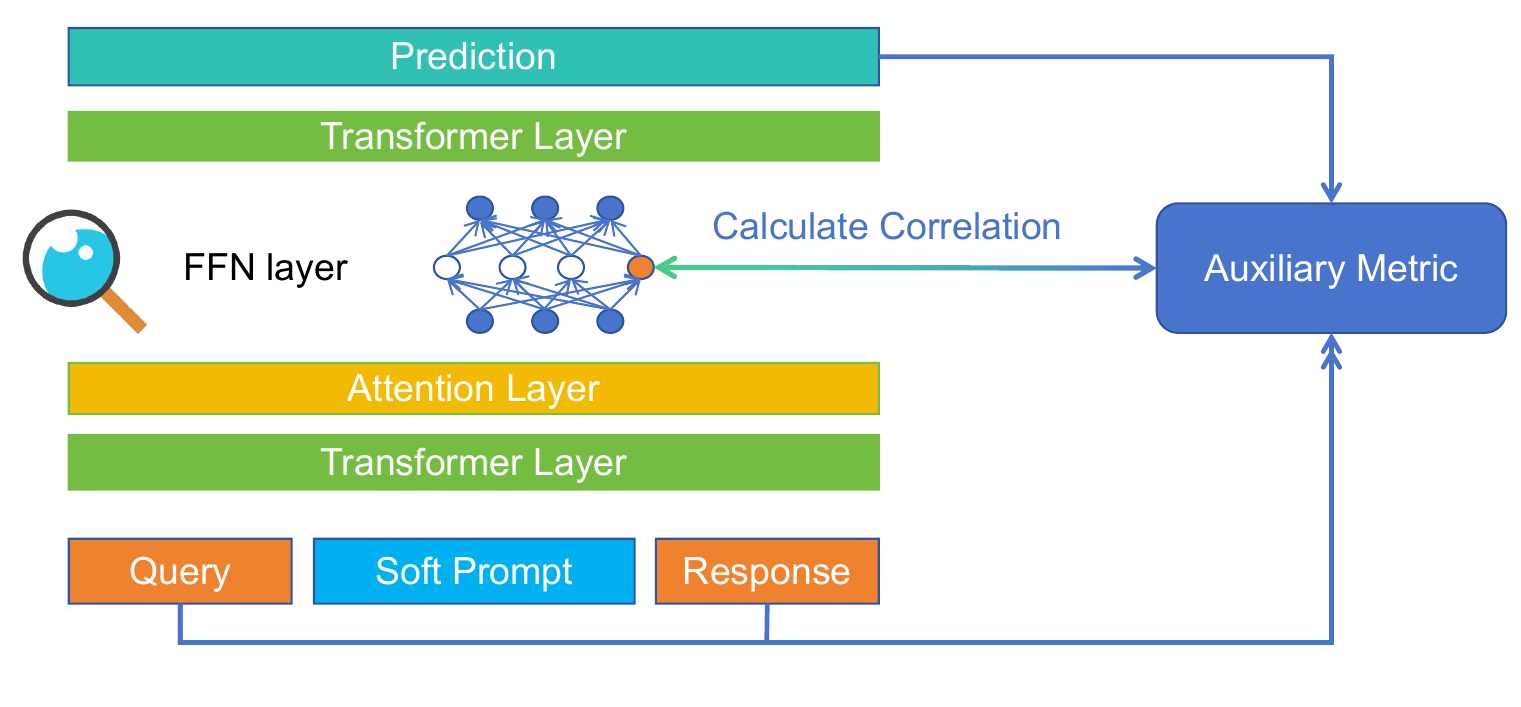}
    \caption{\textbf{Overview of Our Methodology} We calculate the correlation between feedforward-layer neuron activations on a trained soft prompt and an auxiliary metric to identify skill-related neurons.}
    \vspace{-0.2in}
    \label{fig:method}
\end{figure}

In this work, we introduce a simple, lightweight, and broadly applicable method for probing large language models during the observation phase, with a particular focus on identifying neurons associated with specific skills using a downstream corpus. Building on earlier work that isolated “skill neurons” through soft prompt training \citep{lester2021powerscaleparameterefficientprompt,wang2022findingskillneuronspretrained}, our approach expands the analysis from single-skill classification tasks to more complex scenarios involving multiple skills. After training soft prompts on our target tasks, we leverage \textbf{auxiliary metrics}—such as auxiliary labels or the model’s own confidence in its predictions to pinpoint neurons whose activations on the soft prompt are strongly correlated with these metrics corresponding to model skills.

Our framework is also closely related to network dissection, a technique widely used to uncover low-level features in computer vision. Seminal work by \citet{bau2017networkdissectionquantifyinginterpretability} showed how individual neurons in convolutional networks can be linked to specific visual concepts by analyzing their activations in response to corresponding labels. In the NLP domain, similar methods have been employed to identify neurons that capture low-level linguistic phenomena. However, these approaches often struggle with complex, multi-token features because aggregating activations across tokens is not straightforward \citep{gurnee2023findingneuronshaystackcase}. By calculating the activation on the trained soft prompt, we eliminate the need for manual aggregation, thereby better accommodating the inherent complexity of language. Moreover, soft prompt training taps into the full potential of the pretrained model, enabling us to uncover higher-level, task-specific skills encoded within the network.

We empirically demonstrate that our methods can effectively detect neurons with specific skill in open-ended generation (Skill-Mix, \citet{yu2023skillmixflexibleexpandablefamily}) and natural language inference (Heuristic Analysis for NLI Systems, \citet{mccoy2019rightwrongreasonsdiagnosing}). Furthermore, the framework detects neurons corresponding to a previously unknown shortcut in the arithmetic subset in BigBench~\citep{srivastava2023imitationgamequantifyingextrapolating} without reliance on any auxiliary label.

\section{Methodology}


\subsection{Preliminary}
\label{sec:pre}

\paragraph{Prompt Tuning.} Soft prompt tuning \citep{lester2021powerscaleparameterefficientprompt,li-liang-2021-prefix} has been widely adopted to adapt language models to downstream tasks. For a pretrained decoder-based language model $\model$, given an input instruction $\inputsentence$, the embedding function $\embed$ of $\model$ will first maps these tokens to a sequence of $d$-dimensional vectors $\embed(\inputsentence)$, the vectors will then be mapped to a distribution over the vocabulary set, which we will denote as $\Pr_{\model}( \ldots | \embed(\inputsentence))$. Given a training set of instruction $x$ and desired completion $y$, prompt tuning will train $l$ randomly initialized vectors $\{ p_1, \ldots, p_l \}$ to minimize the following loss:

\begin{tiny}
\begin{align*}
    - \E\left[ \frac{1}{|y|} \sum_{j = 1}^{|y|} \log \Pr_{\model}(y_{j} \mid \embed(x), p_1, \ldots p_l, \embed(y_{1:j - 1}))  \right].
\end{align*}
\end{tiny}

We will interpret the pretrained model based on the neurons' activations on the trained prompt. In order to do so, we will (1) freeze the parameter of $\model$ and only train the prompt, and (2) put the soft prompt after the instruction such that the soft prompt can attend to the input instruction through the causal attention mechanism.

\paragraph{Neurons.} Transformer~\citep{vaswani2023attentionneed} is the most common architecture used in language models.
Each layer of a Transformer composed of one attention module and one feed-forward layers. Feed-forward layers in Transformers are typically a two-layer fully connected network operates independently on the hidden state of each token. In the standard LLaMA architecture, the feedforward layer of width $m$ is defined as,

\begin{small}
\begin{align*}
\text{FFN}(h) = W^{(3)} \bigl((W^{(1)} h) \odot \text{SiLU}(W^{(2)} h)\bigr),
\end{align*}
\end{small}

where \(h \in \R^d\) is the input hidden state, \(W^{(1)}, W^{(2)} \in \R^{m \times d},W^{(3)}  \in \R^{d \times m}\) are learnable parameters, \(\odot\) denotes element-wise multiplication, and \(\text{SiLU}(x) = \frac{x}{1 + e^{-x}}\) is the SiLU activation function. We will follow the definition in~\citet{wang2022findingskillneuronspretrained} to define the $i$-th neuron as the set of parameters $\{W^{(1)}_{i, :}, W^{(2)}_{i, :}, W^{(3)}_{:, i}\}$ for $i \in [m]$. The activation of the $i$-th neuron on hidden state $h$ is then defined as $W^{(1)}_{i.:}h \text{SiLU}(W^{(2)}_{i, :} h)$. 

Because the input of the Transformer is always a sequence, each neuron will have different activations at different position of the sequence. We will focus on the activations on the trained soft prompt and use $\activation_{l, i, k}(x)$ to denote the activation on the $k$-th position of the soft prompt for neuron $i$ on layer $l$ when the input instruction is $x$.

\subsection{Method}
\label{sec:method}

We introduces a systematic approach to identifying neurons in language models responsible for activating specific skills in response to tasks. 
Given a training set $S_{\mathrm{train}}$ and a validation set $S_{\mathrm{val}}$ consisting of instruction completion pairs. The method consists of the following stages:

\begin{enumerate}[leftmargin=*]
    \item \textbf{Training:} We will first train soft prompts using a frozen pretrained language model on $S_{\mathrm{train}}$. This enables task-specific adaptation without altering the underlying model weights.
    \item \textbf{Metric Calculation:} After training, we will select a \emph{helper metric} $m$ for each sample in validation set, which is a function of the input sequence and the models' output that maps to a real number. For example, we can map all the sequence to their corresponding loss values given the trained model. We will use $m(S_{\mathrm{val}})$ to denote the calculated metrics. 
    \item \textbf{Neuron Selection:} For every neuron $i$ on layer $l$, we will calculate their activations on the soft prompt over the instruction and completion of the validation set, we will denote the activations on the $k$-th position of the soft prompt as $a_{l, i, k}(S_{\mathrm{val}})$. We then compute the Pearson correlation coefficient between \( a_{l, i, k}(S_{\mathrm{val}}) \) and \( m(S_{\mathrm{val}}) \), defined as:
    
\begin{tiny}
    \begin{align*}
        \mathrm{corr}_{l, i, k}  =  \frac{\sum_{k=1}^{N} \left(a_{l,i,k} - \overline{a_{l,i,k}}\right) \left(m_k - \overline{m}\right)}{\sqrt{\sum_{k=1}^{N} \left(a_{l,i,k} - \overline{a_{l,i,k}}\right)^2} \sqrt{\sum_{k=1}^{N} \left(m_k - \overline{m}\right)^2}}
\end{align*}
 \end{tiny}

\begin{itemize}[leftmargin=*]
    \item $N$ is the number of samples in $S_{\mathrm{val}}$.
    \item $\overline{a_{l,i,k}}$ is the mean activation of neuron $i$ in layer $l$ on the $k$-th position of the soft prompt across the validation set.
    \item $\overline{m}$ is the mean of the helper metric across the validation set.
\end{itemize}

We will then define the correlation of a neuron $l,i$ as $\mathrm{corr}_{l, i} = \max_{k} \mathrm{corr}_{l, i, k}$. 
We then identify neurons with top-$K$ absolute values of correlations where $K$ is a hyperparameter.
    \item \textbf{Interpretation:} We search for sentences in the validation set that maximally or minimally activate the identified neurons. These sentences provide interpretable evidence of potential skills associated with the neuron activations.
\end{enumerate}

We would note that our method is a strict generalization of the method in~\citet{wang2022findingskillneuronspretrained}. In the binary classification setting, if we choose the metric $m$ to map instances of one class to $0$ and instances of the other class to $1$, then we would select neurons whose activations is predictable of the class labels.

\section{Experiments}

We evaluate our framework on the Qwen 1.5 family of instruction-tuned models~\citep{bai2023qwentechnicalreport} (with parameter scale 1.8B) and train soft prompts with $20$ soft tokens using AdamW optimizer with learning rate 3e-3. Our experiments are designed to address the following research questions:

\begin{itemize}
    \item \textbf{RQ1:} Can we detect skill-related neurons in a natural language generation task when provided with explicit meta labels?
    \item \textbf{RQ2:} Can our framework effectively disentangle and isolate neurons that are specialized for fine-grained linguistic cues or heuristics within a uniform task setting?
    \item \textbf{RQ3:} Can our framework detect skill neurons without relying on explicit meta-labels?
\end{itemize}

\paragraph{Skill-Mix (RQ1).}  
To answer RQ1, we build upon the prompting framework introduced by \citet{yu2023skillmixflexibleexpandablefamily}, which guides LLMs to generate natural language sequences that require specific linguistic skills. For instance, the logical skill \emph{spatial reasoning} is defined as the capacity to reason about spatial relationships between objects. We modify this framework to generate question-answer pairs that target one of two skills—spatial reasoning or creating metaphor. An example is shown below:

\begin{tcolorbox}[title=Example Data]
Q: Where is the ball if it is to the right of the box and the box is on the table? \\
A: The ball is to the right of the table.
\end{tcolorbox}

\begin{figure}[h]
    \centering
    \includegraphics[width=1\linewidth]{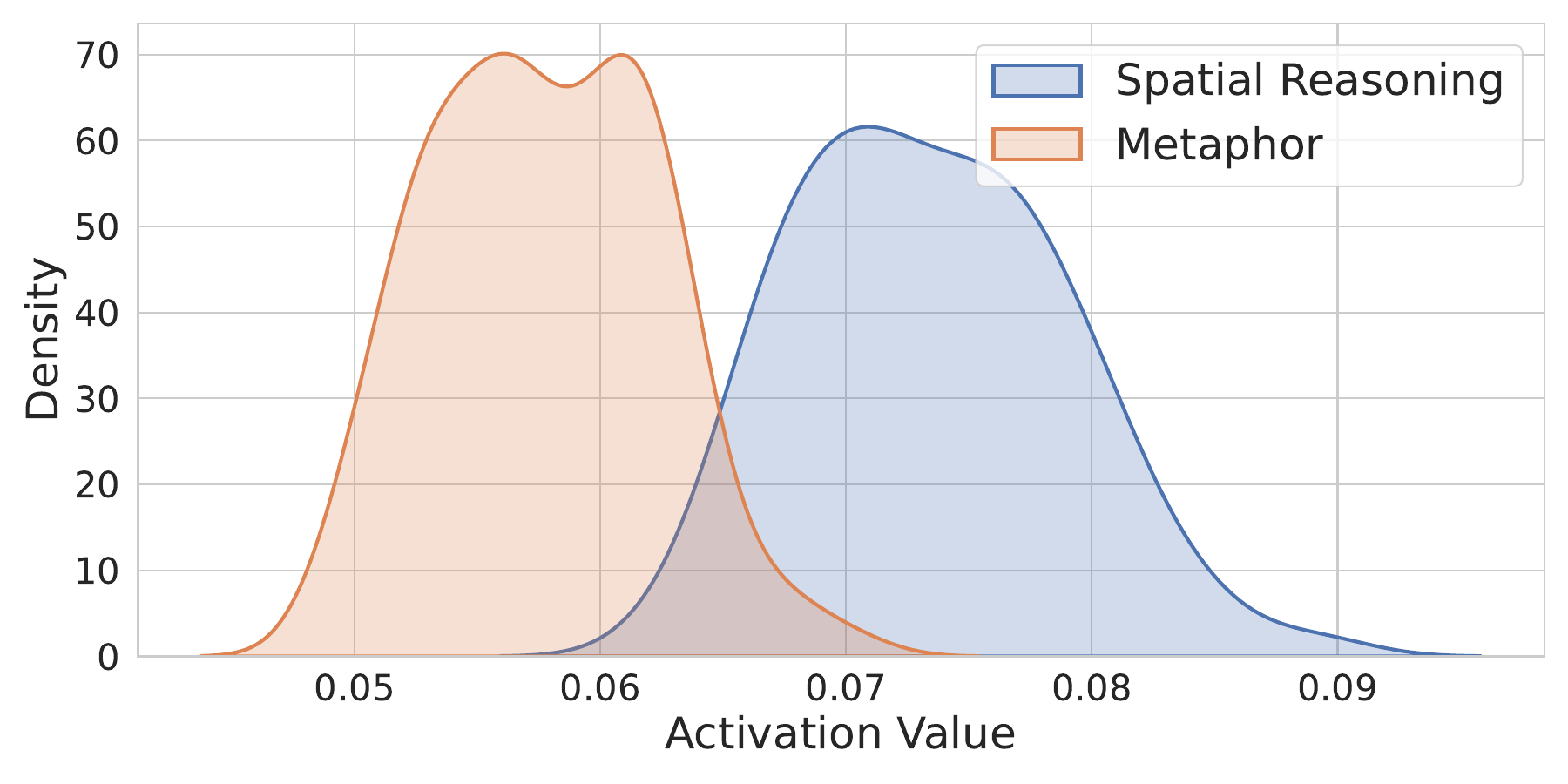}
    \caption{Distribution of activations of the neuron with highest absolute correlation by skill (Skill-Mix). The distribution is interpolated by Kernel density estimation (KDE) based on the empirical distribution of activation over validation set.}
    \label{fig:activation_skillmix}
\end{figure}

We explicitly prompt GPT-4 to produce pairs that leverage one of these two skills and define the metric function $m$ to check whether the selected skill is spatial reasoning. Our analysis reveals a group of neurons with distinct activation patterns corresponding to different skills (\Cref{fig:activation_skillmix}). This experiment demonstrates that the skill differences within LLMs can be reflected in sparse neurons, and our method can well detect these crucial neurons when meta labels are available.

\begin{figure}[h]
    \centering
    \includegraphics[width=1\linewidth]{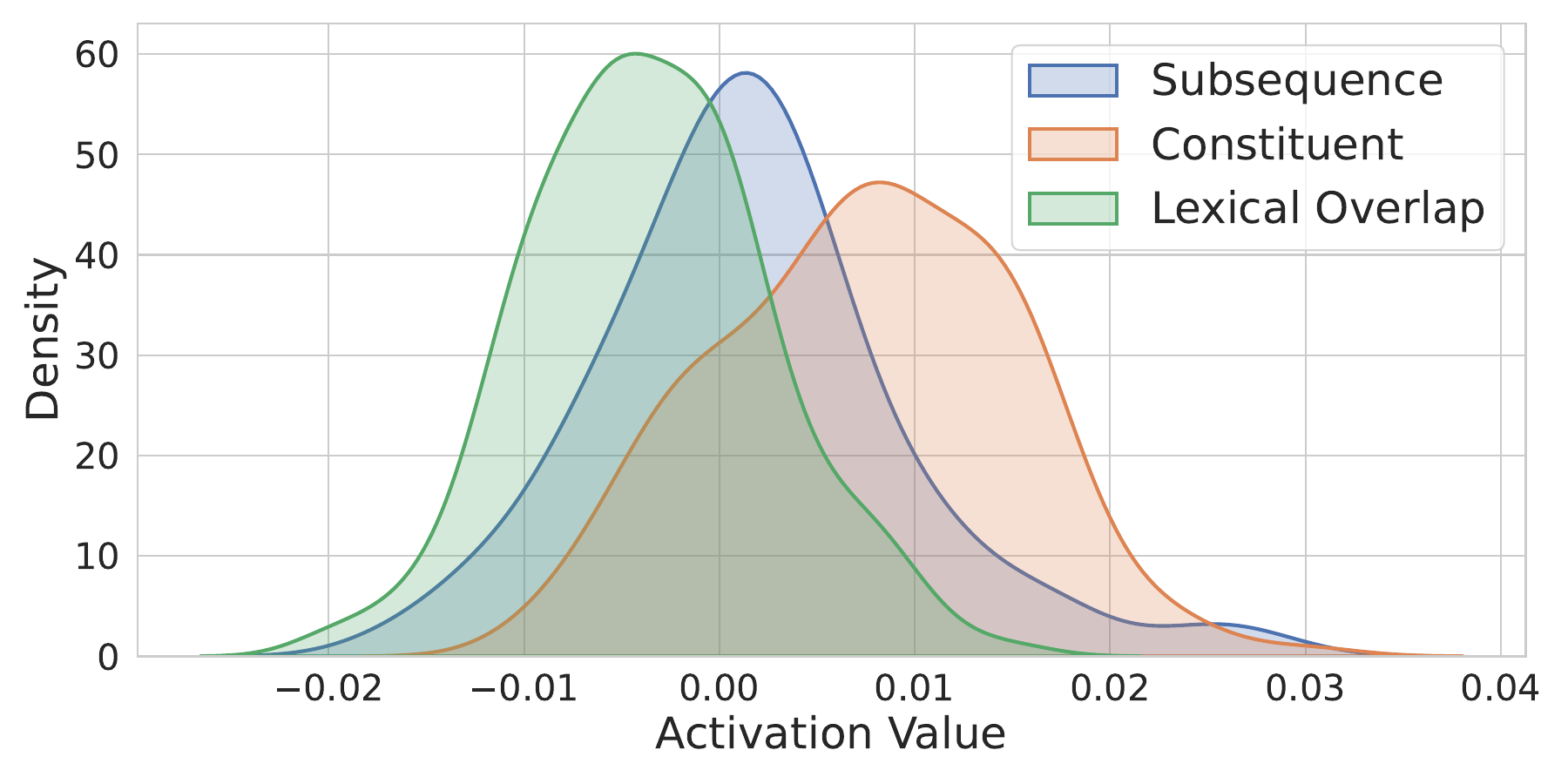}
    \caption{Distribution of activations of the neuron with highest absolute correlation on data with three different heuristics on HANS.}
    \label{fig:hans}
\end{figure}

\paragraph{Heuristic Analysis for NLI Systems (HANS) (RQ2).}  
To investigate RQ2, we apply our framework to the HANS dataset~\citep{mccoy2019rightwrongreasonsdiagnosing}, which is specifically designed to reveal whether LLMs rely on simple syntactic shortcuts for natural language inference. Although the overall task remains natural language inference, the dataset contains various fine-grained heuristics. We define our metric $m$ as whether the heuristic is \emph{Lexical Overlap} and probe the neurons accordingly. As shown in \Cref{fig:hans}, the activation distributions of the chosen neurons vary clearly between different heuristics—even for those heuristics that were not directly used during probing. This result confirms that our framework can identify fine-grained linguistic skill neurons within a single-purpose task.

We further show that high-correlation neurons are sparse on the HANS task. \Cref{fig:corr} illustrates the distribution of correlation values between each neuron's activation and the HANS heuristic label. While we observe a general clustering of neurons around low correlation values, only a small subset of neurons exhibit substantial correlation, exceeding the threshold marked by the red dashed line (0.43). This sparsity further strengthens the findings from our previous analysis showing that our framework can indeed identify fine-grained linguistic skill neurons even when surrounded by a majority of confounding neurons. We defer other two tasks' correlation distribution to~\Cref{sec:add}.

\begin{figure}[h] \centering \includegraphics[width=1\linewidth]{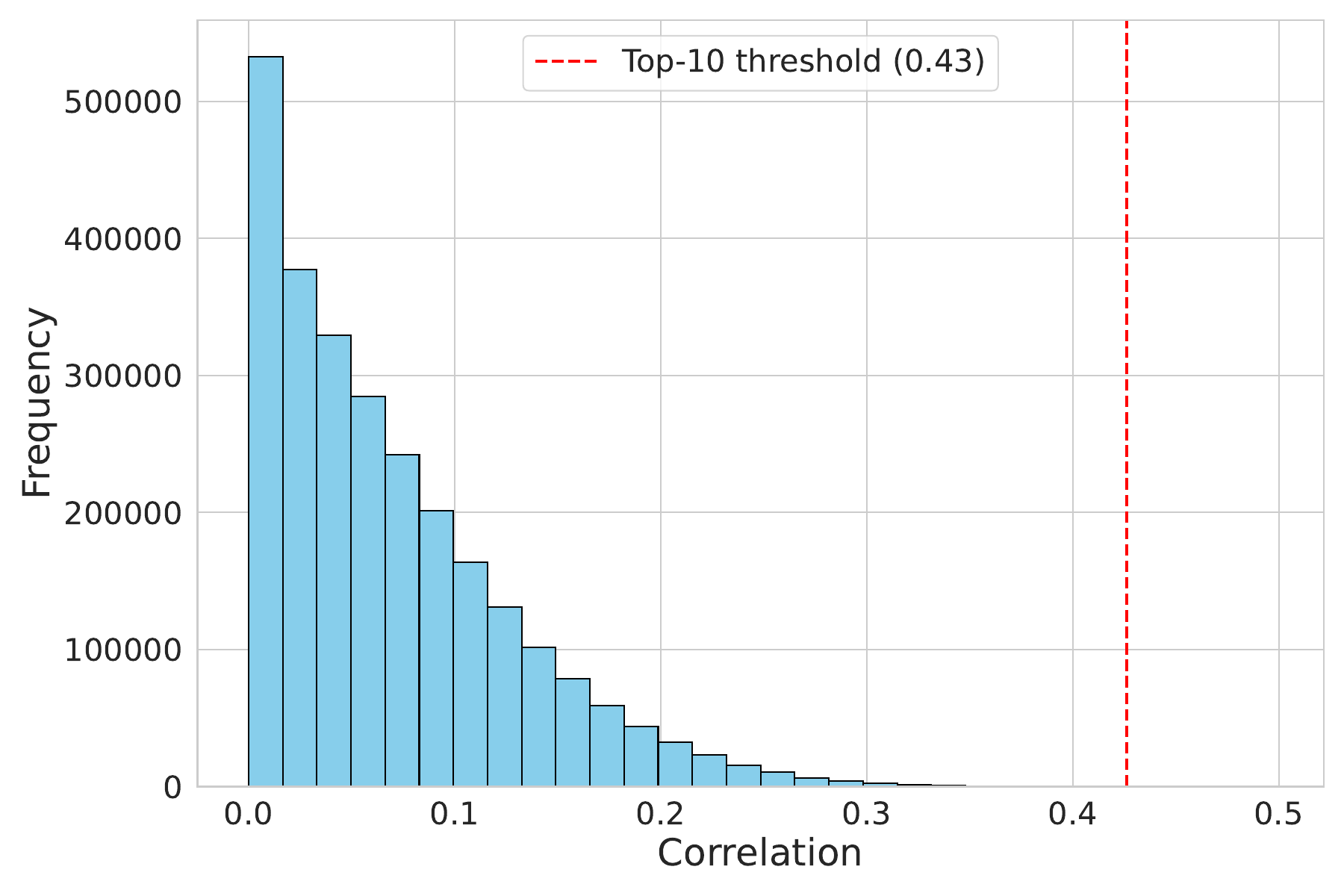} \caption{Distribution of correlation values between each neuron's activation and the HANS heuristic label. The red dashed line indicates the top-10 correlation score threshold (0.43).} \label{fig:corr} \end{figure}

\paragraph{Arithmetic Task (RQ3).}  
For RQ3, we extend our analysis to a task that does not rely on explicit meta labels. We consider a multiple-choice arithmetic problem from BigBench~\citep{srivastava2023imitationgamequantifyingextrapolating} and define our metric $m$ as the per-sample loss for each data point. For the selected neuron, we observe that the top 10 sequences with the lowest activations share a common pattern: (1) the question involves multiplication, and (2) the final answer can be determined by considering only the last digit. An example is provided below:

\begin{tcolorbox}[title=Example Data]
Question: What is 56510 times 52373? \\
Choices: 16619555, 204563610029, \ldots, 2959598230 \\
Answer: 2959598230
\end{tcolorbox}

\begin{figure}[h]
    \centering
    \includegraphics[width=1\linewidth]{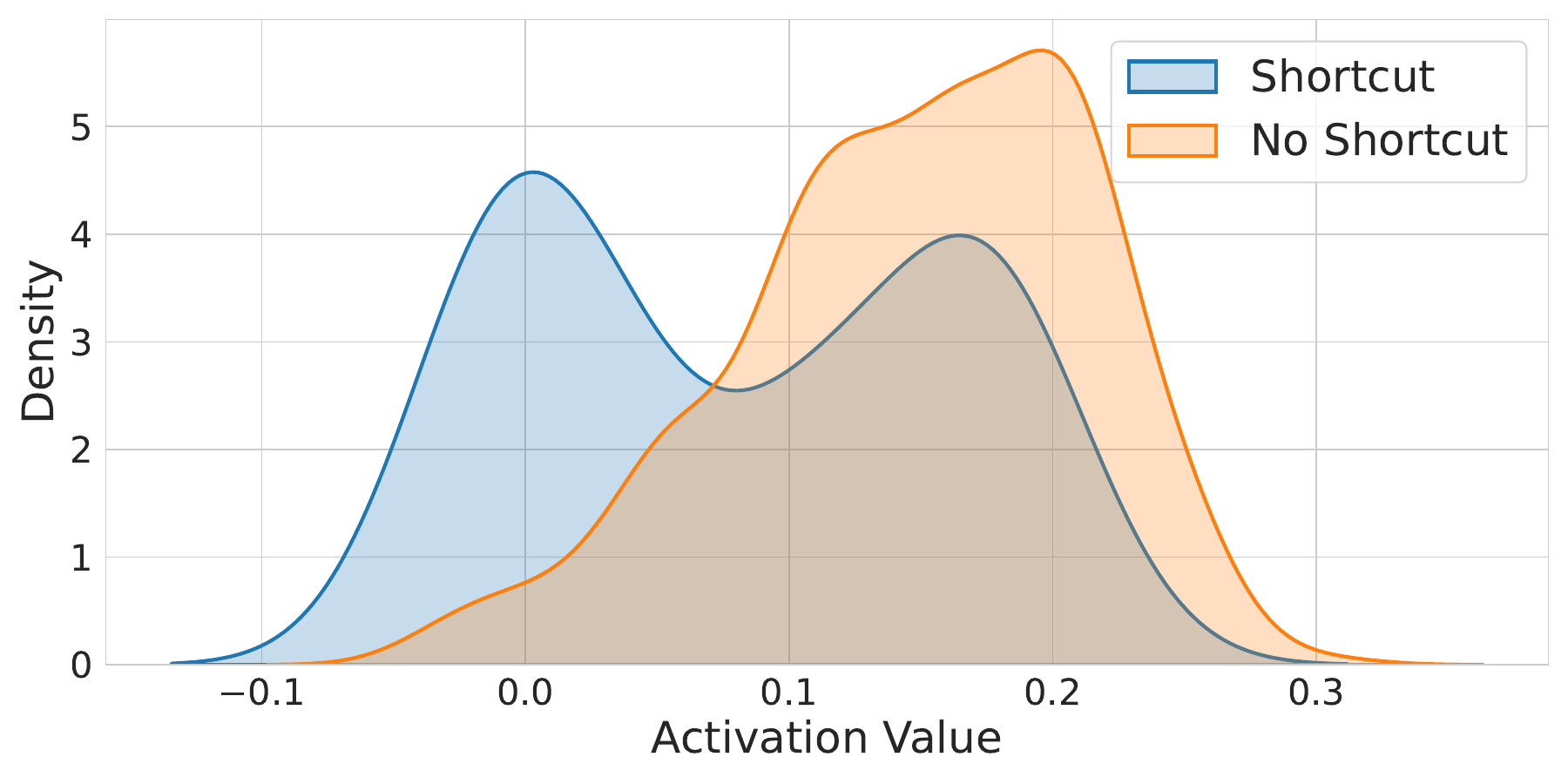}
    \caption{Distribution of activations of the neuron with highest absolute correlation on the data of the Arithmetic task. The shortcut indicates that the correct answer for a multiplication question can be determined solely by the last digit, a pattern automatically discovered by our algorithm.}
    \label{fig:arithmetic}
\end{figure}

We further verify that the selected neuron corresponds to this subskill by plotting the activation distributions separately for data with and without the identified shortcut (\Cref{fig:arithmetic}). The clear distinction observed between these two groups validates that our framework can detect skill neurons even in tasks lacking explicit meta labels.

\section{Conclusion and Future Work}

In this paper, we extend the framework of \citet{wang2022findingskillneuronspretrained} on identifying sparse skill-related neurons within LLMs by examining the correlations between neuron activations and a broad range of model behaviors. 
Experiments on tasks like open-ended question answering, natural language inference, and arithmetic problem-solving demonstrate that our method can consistently identify sparse neurons corresponding to fine-grained linguistic skills and previously unknown heuristics in problem-solving. These findings help understand the skill specialization within LLMs and we encourage future work exploring the causal influence of these identified neurons to model behaviors.

\section{Limitation}

\looseness=-1 Our method, while effective in uncovering task-specific neuron activations via soft prompt training, depends on the quality of prompt tuning and the availability of clear auxiliary metrics. Moreover, the framework and experiments in this paper are designed to discover the internal activation signatures that are highly \textit{correlational} to model behaviors of interest, and we leave examining the \textit{causality} of these signatures to model behaviors to future work.

\bibliography{custom}

@misc{radford2017learninggeneratereviewsdiscovering,
      title={Learning to Generate Reviews and Discovering Sentiment}, 
      author={Alec Radford and Rafal Jozefowicz and Ilya Sutskever},
      year={2017},
      eprint={1704.01444},
      archivePrefix={arXiv},
      primaryClass={cs.LG},
      url={https://arxiv.org/abs/1704.01444}, 
}

@misc{bau2017networkdissectionquantifyinginterpretability,
      title={Network Dissection: Quantifying Interpretability of Deep Visual Representations}, 
      author={David Bau and Bolei Zhou and Aditya Khosla and Aude Oliva and Antonio Torralba},
      year={2017},
      eprint={1704.05796},
      archivePrefix={arXiv},
      primaryClass={cs.CV},
      url={https://arxiv.org/abs/1704.05796}, 
}

@misc{räuker2023transparentaisurveyinterpreting,
      title={Toward Transparent AI: A Survey on Interpreting the Inner Structures of Deep Neural Networks}, 
      author={Tilman Räuker and Anson Ho and Stephen Casper and Dylan Hadfield-Menell},
      year={2023},
      eprint={2207.13243},
      archivePrefix={arXiv},
      primaryClass={cs.LG},
      url={https://arxiv.org/abs/2207.13243}, 
}

@misc{bereska2024mechanisticinterpretabilityaisafety,
      title={Mechanistic Interpretability for AI Safety -- A Review}, 
      author={Leonard Bereska and Efstratios Gavves},
      year={2024},
      eprint={2404.14082},
      archivePrefix={arXiv},
      primaryClass={cs.AI},
      url={https://arxiv.org/abs/2404.14082}, 
}

@misc{vaswani2023attentionneed,
      title={Attention Is All You Need}, 
      author={Ashish Vaswani and Noam Shazeer and Niki Parmar and Jakob Uszkoreit and Llion Jones and Aidan N. Gomez and Lukasz Kaiser and Illia Polosukhin},
      year={2023},
      eprint={1706.03762},
      archivePrefix={arXiv},
      primaryClass={cs.CL},
      url={https://arxiv.org/abs/1706.03762}, 
}

@misc{srivastava2023imitationgamequantifyingextrapolating,
      title={Beyond the Imitation Game: Quantifying and extrapolating the capabilities of language models}, 
      author={Aarohi Srivastava and Abhinav Rastogi and Abhishek Rao and Abu Awal Md Shoeb and Abubakar Abid and Adam Fisch and Adam R. Brown and Adam Santoro and Aditya Gupta and Adrià Garriga-Alonso and Agnieszka Kluska and Aitor Lewkowycz and Akshat Agarwal and Alethea Power and Alex Ray and Alex Warstadt and Alexander W. Kocurek and Ali Safaya and Ali Tazarv and Alice Xiang and Alicia Parrish and Allen Nie and Aman Hussain and Amanda Askell and Amanda Dsouza and Ambrose Slone and Ameet Rahane and Anantharaman S. Iyer and Anders Andreassen and Andrea Madotto and Andrea Santilli and Andreas Stuhlmüller and Andrew Dai and Andrew La and Andrew Lampinen and Andy Zou and Angela Jiang and Angelica Chen and Anh Vuong and Animesh Gupta and Anna Gottardi and Antonio Norelli and Anu Venkatesh and Arash Gholamidavoodi and Arfa Tabassum and Arul Menezes and Arun Kirubarajan and Asher Mullokandov and Ashish Sabharwal and Austin Herrick and Avia Efrat and Aykut Erdem and Ayla Karakaş and B. Ryan Roberts and Bao Sheng Loe and Barret Zoph and Bartłomiej Bojanowski and Batuhan Özyurt and Behnam Hedayatnia and Behnam Neyshabur and Benjamin Inden and Benno Stein and Berk Ekmekci and Bill Yuchen Lin and Blake Howald and Bryan Orinion and Cameron Diao and Cameron Dour and Catherine Stinson and Cedrick Argueta and César Ferri Ramírez and Chandan Singh and Charles Rathkopf and Chenlin Meng and Chitta Baral and Chiyu Wu and Chris Callison-Burch and Chris Waites and Christian Voigt and Christopher D. Manning and Christopher Potts and Cindy Ramirez and Clara E. Rivera and Clemencia Siro and Colin Raffel and Courtney Ashcraft and Cristina Garbacea and Damien Sileo and Dan Garrette and Dan Hendrycks and Dan Kilman and Dan Roth and Daniel Freeman and Daniel Khashabi and Daniel Levy and Daniel Moseguí González and Danielle Perszyk and Danny Hernandez and Danqi Chen and Daphne Ippolito and Dar Gilboa and David Dohan and David Drakard and David Jurgens and Debajyoti Datta and Deep Ganguli and Denis Emelin and Denis Kleyko and Deniz Yuret and Derek Chen and Derek Tam and Dieuwke Hupkes and Diganta Misra and Dilyar Buzan and Dimitri Coelho Mollo and Diyi Yang and Dong-Ho Lee and Dylan Schrader and Ekaterina Shutova and Ekin Dogus Cubuk and Elad Segal and Eleanor Hagerman and Elizabeth Barnes and Elizabeth Donoway and Ellie Pavlick and Emanuele Rodola and Emma Lam and Eric Chu and Eric Tang and Erkut Erdem and Ernie Chang and Ethan A. Chi and Ethan Dyer and Ethan Jerzak and Ethan Kim and Eunice Engefu Manyasi and Evgenii Zheltonozhskii and Fanyue Xia and Fatemeh Siar and Fernando Martínez-Plumed and Francesca Happé and Francois Chollet and Frieda Rong and Gaurav Mishra and Genta Indra Winata and Gerard de Melo and Germán Kruszewski and Giambattista Parascandolo and Giorgio Mariani and Gloria Wang and Gonzalo Jaimovitch-López and Gregor Betz and Guy Gur-Ari and Hana Galijasevic and Hannah Kim and Hannah Rashkin and Hannaneh Hajishirzi and Harsh Mehta and Hayden Bogar and Henry Shevlin and Hinrich Schütze and Hiromu Yakura and Hongming Zhang and Hugh Mee Wong and Ian Ng and Isaac Noble and Jaap Jumelet and Jack Geissinger and Jackson Kernion and Jacob Hilton and Jaehoon Lee and Jaime Fernández Fisac and James B. Simon and James Koppel and James Zheng and James Zou and Jan Kocoń and Jana Thompson and Janelle Wingfield and Jared Kaplan and Jarema Radom and Jascha Sohl-Dickstein and Jason Phang and Jason Wei and Jason Yosinski and Jekaterina Novikova and Jelle Bosscher and Jennifer Marsh and Jeremy Kim and Jeroen Taal and Jesse Engel and Jesujoba Alabi and Jiacheng Xu and Jiaming Song and Jillian Tang and Joan Waweru and John Burden and John Miller and John U. Balis and Jonathan Batchelder and Jonathan Berant and Jörg Frohberg and Jos Rozen and Jose Hernandez-Orallo and Joseph Boudeman and Joseph Guerr and Joseph Jones and Joshua B. Tenenbaum and Joshua S. Rule and Joyce Chua and Kamil Kanclerz and Karen Livescu and Karl Krauth and Karthik Gopalakrishnan and Katerina Ignatyeva and Katja Markert and Kaustubh D. Dhole and Kevin Gimpel and Kevin Omondi and Kory Mathewson and Kristen Chiafullo and Ksenia Shkaruta and Kumar Shridhar and Kyle McDonell and Kyle Richardson and Laria Reynolds and Leo Gao and Li Zhang and Liam Dugan and Lianhui Qin and Lidia Contreras-Ochando and Louis-Philippe Morency and Luca Moschella and Lucas Lam and Lucy Noble and Ludwig Schmidt and Luheng He and Luis Oliveros Colón and Luke Metz and Lütfi Kerem Şenel and Maarten Bosma and Maarten Sap and Maartje ter Hoeve and Maheen Farooqi and Manaal Faruqui and Mantas Mazeika and Marco Baturan and Marco Marelli and Marco Maru and Maria Jose Ramírez Quintana and Marie Tolkiehn and Mario Giulianelli and Martha Lewis and Martin Potthast and Matthew L. Leavitt and Matthias Hagen and Mátyás Schubert and Medina Orduna Baitemirova and Melody Arnaud and Melvin McElrath and Michael A. Yee and Michael Cohen and Michael Gu and Michael Ivanitskiy and Michael Starritt and Michael Strube and Michał Swędrowski and Michele Bevilacqua and Michihiro Yasunaga and Mihir Kale and Mike Cain and Mimee Xu and Mirac Suzgun and Mitch Walker and Mo Tiwari and Mohit Bansal and Moin Aminnaseri and Mor Geva and Mozhdeh Gheini and Mukund Varma T and Nanyun Peng and Nathan A. Chi and Nayeon Lee and Neta Gur-Ari Krakover and Nicholas Cameron and Nicholas Roberts and Nick Doiron and Nicole Martinez and Nikita Nangia and Niklas Deckers and Niklas Muennighoff and Nitish Shirish Keskar and Niveditha S. Iyer and Noah Constant and Noah Fiedel and Nuan Wen and Oliver Zhang and Omar Agha and Omar Elbaghdadi and Omer Levy and Owain Evans and Pablo Antonio Moreno Casares and Parth Doshi and Pascale Fung and Paul Pu Liang and Paul Vicol and Pegah Alipoormolabashi and Peiyuan Liao and Percy Liang and Peter Chang and Peter Eckersley and Phu Mon Htut and Pinyu Hwang and Piotr Miłkowski and Piyush Patil and Pouya Pezeshkpour and Priti Oli and Qiaozhu Mei and Qing Lyu and Qinlang Chen and Rabin Banjade and Rachel Etta Rudolph and Raefer Gabriel and Rahel Habacker and Ramon Risco and Raphaël Millière and Rhythm Garg and Richard Barnes and Rif A. Saurous and Riku Arakawa and Robbe Raymaekers and Robert Frank and Rohan Sikand and Roman Novak and Roman Sitelew and Ronan LeBras and Rosanne Liu and Rowan Jacobs and Rui Zhang and Ruslan Salakhutdinov and Ryan Chi and Ryan Lee and Ryan Stovall and Ryan Teehan and Rylan Yang and Sahib Singh and Saif M. Mohammad and Sajant Anand and Sam Dillavou and Sam Shleifer and Sam Wiseman and Samuel Gruetter and Samuel R. Bowman and Samuel S. Schoenholz and Sanghyun Han and Sanjeev Kwatra and Sarah A. Rous and Sarik Ghazarian and Sayan Ghosh and Sean Casey and Sebastian Bischoff and Sebastian Gehrmann and Sebastian Schuster and Sepideh Sadeghi and Shadi Hamdan and Sharon Zhou and Shashank Srivastava and Sherry Shi and Shikhar Singh and Shima Asaadi and Shixiang Shane Gu and Shubh Pachchigar and Shubham Toshniwal and Shyam Upadhyay and Shyamolima and Debnath and Siamak Shakeri and Simon Thormeyer and Simone Melzi and Siva Reddy and Sneha Priscilla Makini and Soo-Hwan Lee and Spencer Torene and Sriharsha Hatwar and Stanislas Dehaene and Stefan Divic and Stefano Ermon and Stella Biderman and Stephanie Lin and Stephen Prasad and Steven T. Piantadosi and Stuart M. Shieber and Summer Misherghi and Svetlana Kiritchenko and Swaroop Mishra and Tal Linzen and Tal Schuster and Tao Li and Tao Yu and Tariq Ali and Tatsu Hashimoto and Te-Lin Wu and Théo Desbordes and Theodore Rothschild and Thomas Phan and Tianle Wang and Tiberius Nkinyili and Timo Schick and Timofei Kornev and Titus Tunduny and Tobias Gerstenberg and Trenton Chang and Trishala Neeraj and Tushar Khot and Tyler Shultz and Uri Shaham and Vedant Misra and Vera Demberg and Victoria Nyamai and Vikas Raunak and Vinay Ramasesh and Vinay Uday Prabhu and Vishakh Padmakumar and Vivek Srikumar and William Fedus and William Saunders and William Zhang and Wout Vossen and Xiang Ren and Xiaoyu Tong and Xinran Zhao and Xinyi Wu and Xudong Shen and Yadollah Yaghoobzadeh and Yair Lakretz and Yangqiu Song and Yasaman Bahri and Yejin Choi and Yichi Yang and Yiding Hao and Yifu Chen and Yonatan Belinkov and Yu Hou and Yufang Hou and Yuntao Bai and Zachary Seid and Zhuoye Zhao and Zijian Wang and Zijie J. Wang and Zirui Wang and Ziyi Wu},
      year={2023},
      eprint={2206.04615},
      archivePrefix={arXiv},
      primaryClass={cs.CL},
      url={https://arxiv.org/abs/2206.04615}, 
}

@misc{mccoy2019rightwrongreasonsdiagnosing,
      title={Right for the Wrong Reasons: Diagnosing Syntactic Heuristics in Natural Language Inference}, 
      author={R. Thomas McCoy and Ellie Pavlick and Tal Linzen},
      year={2019},
      eprint={1902.01007},
      archivePrefix={arXiv},
      primaryClass={cs.CL},
      url={https://arxiv.org/abs/1902.01007}, 
}

@misc{yu2023skillmixflexibleexpandablefamily,
      title={Skill-Mix: a Flexible and Expandable Family of Evaluations for AI models}, 
      author={Dingli Yu and Simran Kaur and Arushi Gupta and Jonah Brown-Cohen and Anirudh Goyal and Sanjeev Arora},
      year={2023},
      eprint={2310.17567},
      archivePrefix={arXiv},
      primaryClass={cs.CL},
      url={https://arxiv.org/abs/2310.17567}, 
}

@misc{bai2023qwentechnicalreport,
      title={Qwen Technical Report}, 
      author={Jinze Bai and Shuai Bai and Yunfei Chu and Zeyu Cui and Kai Dang and Xiaodong Deng and Yang Fan and Wenbin Ge and Yu Han and Fei Huang and Binyuan Hui and Luo Ji and Mei Li and Junyang Lin and Runji Lin and Dayiheng Liu and Gao Liu and Chengqiang Lu and Keming Lu and Jianxin Ma and Rui Men and Xingzhang Ren and Xuancheng Ren and Chuanqi Tan and Sinan Tan and Jianhong Tu and Peng Wang and Shijie Wang and Wei Wang and Shengguang Wu and Benfeng Xu and Jin Xu and An Yang and Hao Yang and Jian Yang and Shusheng Yang and Yang Yao and Bowen Yu and Hongyi Yuan and Zheng Yuan and Jianwei Zhang and Xingxuan Zhang and Yichang Zhang and Zhenru Zhang and Chang Zhou and Jingren Zhou and Xiaohuan Zhou and Tianhang Zhu},
      year={2023},
      eprint={2309.16609},
      archivePrefix={arXiv},
      primaryClass={cs.CL},
      url={https://arxiv.org/abs/2309.16609}, 
}

@inproceedings{li-liang-2021-prefix,
    title = "Prefix-Tuning: Optimizing Continuous Prompts for Generation",
    author = "Li, Xiang Lisa  and
      Liang, Percy",
    editor = "Zong, Chengqing  and
      Xia, Fei  and
      Li, Wenjie  and
      Navigli, Roberto",
    booktitle = "Proceedings of the 59th Annual Meeting of the Association for Computational Linguistics and the 11th International Joint Conference on Natural Language Processing (Volume 1: Long Papers)",
    month = aug,
    year = "2021",
    address = "Online",
    publisher = "Association for Computational Linguistics",
    url = "https://aclanthology.org/2021.acl-long.353",
    doi = "10.18653/v1/2021.acl-long.353",
    pages = "4582--4597",
}

@misc{bills2023language,
 title={Language models can explain neurons in language models},
 author={
    Bills, Steven and Cammarata, Nick and Mossing, Dan and Tillman, Henk and Gao, Leo and Goh, Gabriel and Sutskever, Ilya and Leike, Jan and Wu, Jeff and Saunders, William
 },
 year={2023},
 howpublished = {\url{https://openaipublic.blob.core.windows.net/neuron-explainer/paper/index.html}}
}

@misc{choi2024automatic,
  author       = {Choi, Dami and Huang, Vincent and Meng, Kevin and Johnson, Daniel D and Steinhardt, Jacob and Schwettmann, Sarah},
  title        = {Scaling Automatic Neuron Description},
  year         = {2024},
  month        = {October},
  day          = {23},
  howpublished = {\url{https://transluce.org/neuron-descriptions}}
}

@misc{lester2021powerscaleparameterefficientprompt,
      title={The Power of Scale for Parameter-Efficient Prompt Tuning}, 
      author={Brian Lester and Rami Al-Rfou and Noah Constant},
      year={2021},
      eprint={2104.08691},
      archivePrefix={arXiv},
      primaryClass={cs.CL},
      url={https://arxiv.org/abs/2104.08691}, 
}

@misc{gurnee2023findingneuronshaystackcase,
      title={Finding Neurons in a Haystack: Case Studies with Sparse Probing}, 
      author={Wes Gurnee and Neel Nanda and Matthew Pauly and Katherine Harvey and Dmitrii Troitskii and Dimitris Bertsimas},
      year={2023},
      eprint={2305.01610},
      archivePrefix={arXiv},
      primaryClass={cs.LG},
      url={https://arxiv.org/abs/2305.01610}, 
}

@misc{wang2022findingskillneuronspretrained,
      title={Finding Skill Neurons in Pre-trained Transformer-based Language Models}, 
      author={Xiaozhi Wang and Kaiyue Wen and Zhengyan Zhang and Lei Hou and Zhiyuan Liu and Juanzi Li},
      year={2022},
      eprint={2211.07349},
      archivePrefix={arXiv},
      primaryClass={cs.CL},
      url={https://arxiv.org/abs/2211.07349}, 
}

\appendix
\newpage
\section{Additional Experiments}
\label{sec:add}

We show the distribution of correlation scores of neurons for SkillMix and Arithmetic tasks here. We observe that the selected neurons' activation has high correlations with the auxiliary metrics and only a very sparse subset of neurons has the same level of correlations.

\begin{figure}[h] \centering \includegraphics[width=1\linewidth]{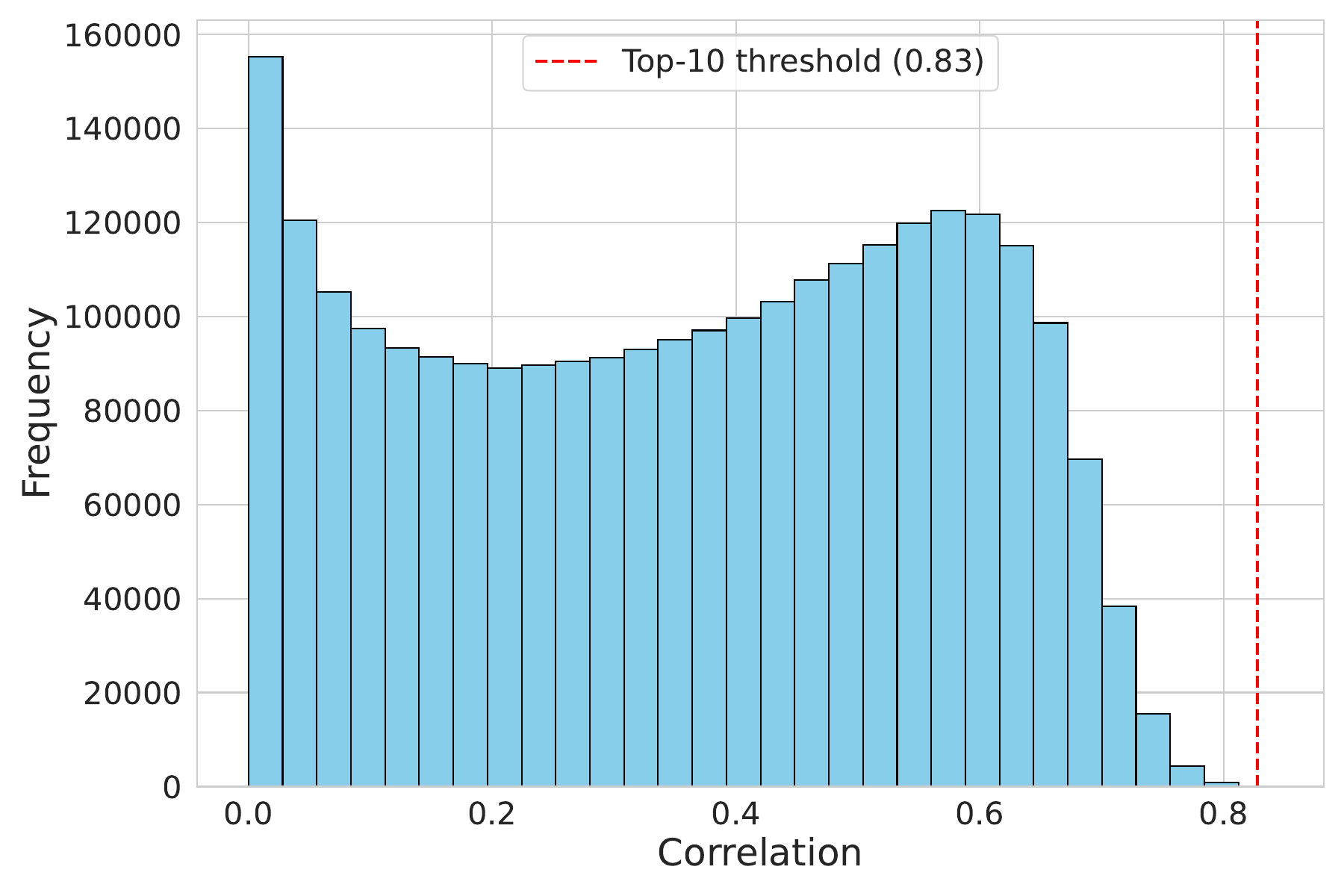} \caption{Distribution of correlation values between each neuron's activation and the skill label (for the 1.4B model). The red dashed line indicates the top-10 threshold (0.83).} \label{fig:corr} \end{figure}

\begin{figure}[h] \centering \includegraphics[width=1\linewidth]{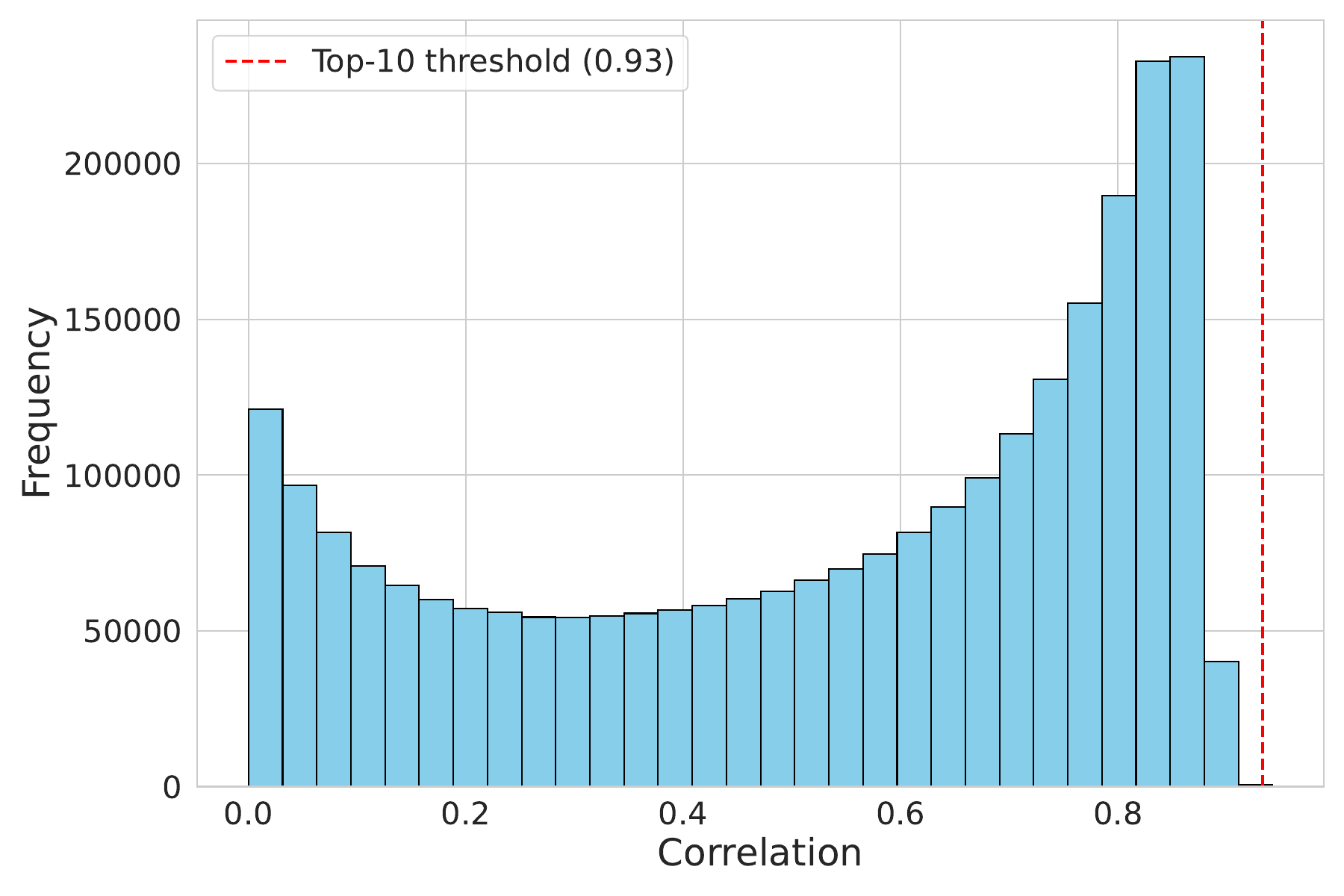} \caption{Distribution of correlation values between each neuron's activation and the arithmetic validation sample loss (for the 1.4B model). The red dashed line indicates the top-10 threshold (0.93).} \label{fig:corr} \end{figure}

\end{document}